\documentclass{article}

\usepackage[final]{corl_2020} 
\usepackage{graphicx}
\usepackage{svg}
\usepackage{amsmath,amsfonts,amssymb}
\usepackage{enumitem}
 \pdfoutput=1

\title{Learning Vision-based Reactive Policies for Obstacle Avoidance}

\author{
  Elie Aljalbout, Ji Chen, Konstantin Ritt, Maximilian Ulmer, and Sami Haddadin\\
  Munich School of Robotics and Machine Intelligence\\
  Technical University of Munich\\
  \texttt{\{name.lastname@tum.de\}} 
}

\begin{document}
\maketitle


\begin{abstract}
    In this paper, we address the problem of vision-based obstacle avoidance for robotic manipulators. This topic poses challenges for both perception and motion generation. While most work in the field aims at improving one of those aspects, we provide a unified framework for approaching this problem. The main goal of this framework is to connect perception and motion by identifying the relationship between the visual input and the corresponding motion representation. To this end, we propose a method for learning reactive obstacle avoidance policies. We evaluate our method on goal-reaching tasks for single and multiple obstacles scenarios. We show the ability of the proposed method to efficiently learn stable obstacle avoidance strategies at a high success rate, while maintaining closed-loop responsiveness required for critical applications like human-robot interaction.
\end{abstract}



\section{Introduction}

During task execution, robots should be capable of operating in their workspaces without colliding with obstacles. In well-structured environments, this constraint can be easily ensured by carefully designing collision-free motion trajectories based on the understanding of the robot surroundings. In contrast, unstructured environments present the challenge of autonomously reacting to previously unknown settings. To tackle this challenge, extra efforts are needed in order to design proper perception systems, capable of understanding the environment, as well as reactive strategies to avoid the obstacles. In this work, we are concerned with obstacle avoidance for robot manipulators. In addition to the previously mentioned challenges, such systems impose additional constraints such as joint limits, singularities and self-collision. All of these aspects add to the complexity of the problem, and require proper care in the formulation of both classical and learning-based methods.

In this context, proprioceptive robot sensors enable collision detection and early reactions which can prevent substantial damages to the robot and its environment~\citep{haddadin2008collision}. On the other hand, exteroceptive sensing (e.g. cameras) provides geometric information and objects poses which allow when required, to avoid the obstacles in the first place. This motivates vision-based motion generation and planning for obstacle avoidance. Although this problem seems trivial at first glance, it entails a multitude of challenges. Namely, most approaches in this field are based on the classical perception and planning pipeline~\citep{kappler2018real, rai2017learning}. Every module in such a pipeline introduces extra errors to the overall system. Namely, state-of-the-art pose detection methods (such as DeepIM~\citep{li2018deepim} and Pix2Pose~\citep{park2019pix2pose}) still lack the required accuracy for such tasks. Similarly, real-time generation of reactive motion trajectories could be error prone, especially when the given obstacle poses are faulty.

One way to avoid the aforementioned problems is to learn model-free reactive motion policies. This can be simply done using end-to-end deep reinforcement learning (DRL). However, these methods could lack the required generalization capabilities which defies the original purpose. Additionally, DRL algorithms require long interaction times, which are time and resource expensive and could also be damaging to the robot, unless safe exploration is guaranteed~\citep{pecka2014safe, zhu2020ingredients}. Furthermore, with the lack of proper action space design, DRL methods tend to fail in real-world robot learning tasks~\citep{martin2019variable}. Instead, it is desirable to combine DRL with classical methods as a trade-off between sample-efficiency, interpretability, modelling efforts and modelling errors.

In this work, we adopt the latter approach. Our main goal is to relate the visual information to the robot's motion. To do that, we learn a residual strategy to react to obstacles in the robot's surroundings based on vision. This policy augments another simple hand-crafted one. Both are designed based on Riemannian Motion Policies (RMP)~\citep{ratliff2018riemannian}, which enables us to effectively combine them, while keeping desirable constraints such as joint limits. To further reduce the sample complexity of our method, we learn a visual latent model for images. This model enables us to reduce the complexity of our policies and also to have an interpretable and generalizable representation of the environment. Our contributions can be summarized as follows:
\begin{itemize}[leftmargin=*]
    \item We present a unified framework for learning vision-based obstacle avoidance with minimum expert knowledge.
    \item Our method can learn successful single obstacle avoidance strategies within less than an hour of real-time interactions with the environment, using a single manipulator.
    \item Our method enables simultaneous avoidance of multiple obstacles at a high success rate.
    \item To our knowledge, our method is the first to use DRL for vision-based obstacle avoidance on robot manipulators.
\end{itemize}


\section{Related Work}
\textbf{Obstacle Avoidance} is a fundamental problem in robotics for which many solutions have been proposed. Possible solutions can be categorized into two main groups: local and global methods~\citep{khansari2012dynamical}. Local algorithms, such as potential fields~\citep{khatib1986real}, only adapt the manipulator's behavior in the presence of obstacles. They are reactive closed-loop control techniques which generate local motion. Due to their low complexity, these algorithms are able to run in the inner control loop of the robot~\citep{haddadin2010real} which ensures the level of responsiveness that is needed for critical scenarios, such as safe human-robot interaction. However, these approaches assume Euclidean geometry in the task space of the manipulator~\citep{khatib1987unified} and only use internal geometry of the kinematic chain~\citep{cheng2018rmpflow}. Despite their simplicity and versatility, purely reactive methods often culminate in undesirable behaviors such as instability or oscillation~\citep{ratliff2018riemannian}.
On the other hand, global planning-based, differential geometric approaches aim to model the intrinsically non-euclidean task space of the robot~\citep{ratliff2015understanding}. 
This thread of research has seen significant progress recently and primarily uses optimization to generate nonlinear trajectories~\citep{watterson2018trajectory}. Nonetheless, planning-based method are notoriously computationally expensive and lack the reactiveness of local methods.
To avoid obstacles, these methods require knowledge about obstacles (e.g. keypoints) which is crucial for autonomous applications in unstructured, dynamic environments. This need for semantic understanding of the environment has sprouted a variety of research on vision-based obstacle avoidance, which is especially prominent in mobile robotics~\citep{guzel2011vision}. 

\textbf{Motion Generation.} Movement primitives (MP) are a common and versatile method to encode motion as a sequence of unit actions. Popular examples for this paradigm are Dynamic Movement Primitives (DMP)~\citep{schaal2006dynamic} and  Probabilistic Movement Primitives (ProMP)~\citep{paraschos2013probabilistic}. DMPs encode motion as second-order differential equations. They can scale motion both in space and time, and enable learning by demonstration in their formulation~\citep{schaal2006dynamic}. In contrast, ProMPs encode motion primitives by a distribution over trajectories which allows the use of powerful statistical methods, such as conditioning a certain trajectory on some desired behavior. Similar to RMPs, some motion primitives model motion as a second-order differential equation, however, they are lacking the modularity and combinatorial capabilities of operational space closed-loop control~\citep{khatib1987unified}.

\textbf{Reactive Planning.} Responsiveness is a crucial factor in robotic manipulation tasks with a high degree of uncertainty due to incomplete modelling, inaccurate sensors or incomplete observability~\citep{kappler2018real}. Combining local reactive control and global collision-free motion planning into a unified \textit{reactive planning} framework is common in mobile robotics~\citep{kuwata2009real, abraham2008reactive}, however only a few approaches are specifically designed for high degrees of freedom (DoF) manipulators~\citep{schaal2006dynamic, brock2002elastic, rai2017learning, ratliff2018riemannian}. In this work, we follow the route of a hybrid approach combining the reactive and planning capabilities of RMPs with the flexibility of DRL.

\textbf{Reinforcement Learning} for robotics and continuous control has seen substantial progress in recent years~\citep{bodnar2019quantile}. DRL has the potential to deliver on the promise of closing the loop between perception and motion, without the manual design of intricate features or behaviors~\citep{kober2013reinforcement}. Similar to the case of classical vision-based obstacle avoidance methods, there is a comprehensive body of literature focused on mobile robots and navigation using RL~\citep{kahn2017uncertainty, loquercio2018dronet, lin2020robust}. Only few work has been done in this area when it comes to robot manipulators, however, while keeping vision out of the loop and still assuming explicit a priori knowledge about obstacles~\citep{sangiovanni2018deep}.
To this end, we combine a vision-based reactive policy with a hand-designed, task-specific baseline policy, to leverage the advantages of model-free RL for reactive obstacle avoidance. We train the reactive policy to augment the baseline based on residual RL~\citep{silver2018residual, johannink2019residual}. By combining learning and control, our approach is robust to common pitfalls of reinforcement learning, such as stability, long-horizon and sparse-rewards.

In summary, vision-based obstacle avoidance is a highly researched topic in robotics. For navigation, many approaches have been presented to improve the accuracy, robustness and overall performance~\citep{guzel2011vision,kahn2017uncertainty, lin2020robust}. In general, such methods are not directly transferable to high-DoF manipulators. While the problems seem similar at a high level, they are fundamentally different. In mobile navigation the geometry of the robot is preserved throughout execution. This is not the case for articulated manipulators. This aspect introduces extra concerns such as self-collision, joint-limit avoidance, or various motion and torque/force constraints. Methods focused on high-DoF manipulators assume perfect knowledge about the obstacle position and are based on classical robotic pipelines separating vision and planning \citep{brock2002elastic,ratliff2018riemannian,schaal2006dynamic}, which usually results in reduced performance \citep{levine2016end}, or rely on kinesthetic demonstrations~\cite{rai2017learning}. In contrast, our approach does not make any expensive assumptions nor does it separate vision from planning, rather treats the two in a unified formalism.


\section{Background}
The most salient attributes of robotic manipulators are best expressed in terms of differential geometry~\citep{kimgeometric}. More specifically, motion generation and control can be seen as the problem of transforming desired behavior from one or multiple smooth manifolds -- representing the task space $\mathcal{T}$ -- to another smooth manifold $\mathcal{C}$, the configuration space. These manifolds are related by a differential map $\phi:\mathcal{C}\rightarrow\mathcal{T}$, called the task map. By equipping these manifolds with a Riemannian metric $\textbf{M}$, we can elegantly relate notions like angles and distances to design a curve $q(t) \in \mathcal{C}$ which implements the desired behavior of $\mathcal{T}$. 

\textbf{Riemannian Motion Policies.}
An RMP is a motion representation defined on a Riemannian manifold. The generation of this type of policies depends not only on the potential field existing in the space, but also on the curvature of the space itself~\citep{ratliff2018riemannian}. Consider an arbitrary $m$-dimensional manifold $\mathcal{M}$ with generalized coordinates $\textbf{x}\in\mathbb{R}^m$. In its canonical form, an RMP is defined as $(\textbf{a},\textbf{M})_\mathcal{M}$, where $\textbf{a}:\textbf{x},\dot{\textbf{x}} \mapsto \ddot{\textbf{x}}^d \in\mathbb{R}^m$ represents a second-order dynamical system mapping $\textbf{x}$ and $\dot{\textbf{x}}$ to desired accelerations $\ddot{\textbf{x}}^d$, and $\textbf{M}=\textbf{M}(\textbf{x},\dot{\textbf{x}})\in\mathbb{R}^{m\times m}$ is a Riemannian metric which varies smoothly with the state $(\textbf{x},\dot{\textbf{x}})$. We can interpret $\textbf{M}$ as an inertial matrix which also defines the weight of the RMP when combined with others. In its natural form, an RMP is defined as $(\textbf{f},\textbf{M})_\mathcal{M}$, where $\textbf{f}$ indicates the map from position and velocity to the desired force. The force map $\textbf{f}$ in the natural form and the acceleration map in the canonical form has the relation $\textbf{f}=\textbf{M}\textbf{a}$. 
The natural form of an RMP is commonly used when space transformations are to be applied, due to the lower computational complexity of such operations using this form.

The RMP framework also provides \texttt{push}, \texttt{pull} and \texttt{add} operators. \texttt{push} and \texttt{pull} can transform an RMP defined in a certain task space into another, based on the task map $\phi$ and its Jacobian $\textbf{J}$. 
As for \texttt{add}, it is used to compose RMPs defined in the same task space into one policy.

\textbf{RMPflow}\label{section:RMPflow} is a policy synthesis framework based on~\citep{cheng2018rmpflow}. In this method, RMP-tree is introduced as a tree structured computational graph. Each node in such a tree represents a Riemannian manifold and is equipped with an RMP. The root node of the tree describes the configuration space $\mathcal{C}$ of the robot where the global joint space policy is defined. The leaf nodes represent task spaces, where the designed or learned local policies are defined.

The RMPs in this framework are characterized by a set of geometric dynamical systems (GDSs). Consider a manifold $\mathcal{M}$ with generalized coordinate $\textbf{x}\in\mathbb{R}^m$, a GDS defined on such manifold can be expressed as:
\begin{equation}
    \label{eq:GDS}
    (\textbf{G}(\textbf{x}, \dot{\textbf{x}}) + \Xi_{\textbf{G}}(\textbf{x},\dot{\textbf{x}}))\ddot{\textbf{x}}+\xi_\textbf{G}(\textbf{x},\dot{\textbf{x}}) = -\nabla\Phi(\textbf{x})-\textbf{B}(\textbf{x}, \dot{\textbf{x}})\dot{\textbf{x}},
\end{equation}
where $\textbf{G}: \mathbb{R}^m\times\mathbb{R}^m\rightarrow \mathbb{R}_+^{m\times m}$ is referred to as the metric matrix, $\textbf{B}: \mathbb{R}^m\times\mathbb{R}^m\rightarrow \mathbb{R}_+^{m\times m}$ as the  damping matrix and $\Phi:\mathbb{R}^m\rightarrow \mathbb{R}$ as the potential function. Additionally,    $\Xi_{\textbf{G}} := \frac{1}{2}\sum_{i=1}^m \dot{x}_i\partial_{\dot{\textbf{x}}} \textbf{g}_i$ and $
    \xi_{\textbf{G}} := \overset{\textbf{x}}{\textbf{G}}\dot{\textbf{x}}-\frac{1}{2}\nabla_{\textbf{x}}(\dot{\textbf{x}}^\intercal\textbf{G}\dot{\textbf{x}})$
are curvature terms induced by the metric $\textbf{G}$, where $\textbf{g}_i$ denotes the $i$-th column of $\textbf{G}$ and $\overset{\textbf{x}}{\textbf{G}}:=[\partial_{\textbf{x}}\textbf{g}_i\dot{\textbf{x}}]_{i=1}^m$. Based on the GDS formulation, the RMP metric term $\textbf{M}$ is defined as $\textbf{M}:=\textbf{G}+\Xi_{\textbf{G}}$. The forcing term of RMP can be calculated by moving the curvature term $\xi_{\textbf{G}}$ to the right hand side of equation~\ref{eq:GDS}, we then obtain:
 \begin{equation}
 \label{eq:RMP_force}
     \textbf{f} = (\textbf{M}\ddot{\textbf{x}})=-\xi_{\textbf{G}}-\nabla\Phi-\textbf{B}\dot{\textbf{x}}
 \end{equation}
RMP-algebra introduces the \texttt{pushforward}, \texttt{pullback} and \texttt{resolve} operators used to construct the policy generation process. The goal of this process is to generate the desired accelerations $\ddot{\textbf{q}}^d$ in the root node of the tree (robot configuration space) based on $\textbf{q}$ and $\dot{\textbf{q}}$. 
For further information about this framework, we refer the readers to the original paper~\citep{cheng2018rmpflow}.

\textbf{Reinforcement Learning} (RL) is a computational approach to automate policy learning by  maximizing  cumulative reward in an environment~\citep{sutton2018reinforcement}. RL tasks are usually formulated as Markov Decision Processes (MDP). A finite-horizon, discounted MDP is characterized by the tuple $\mathcal{M} = (\mathcal{S}, \mathcal{A}, \mathcal{P}, r, \rho_0, \gamma, T )$, where the state and action spaces are respectively $\mathcal{S}$ and $\mathcal{A}$ , transition dynamics $\mathcal{P}: \mathcal{S} \times \mathcal{A} \to \mathcal{S}$, reward $r: \mathcal{S} \times \mathcal{A} \to \mathbb{R} $, an initial state distribution $\rho_0$, discount factor $\gamma \in [0,1]$, and horizon $T$. The optimal policy $\pi : \mathcal{S} \to P(\mathcal{A})$, maximizes the expected discounted reward:
\begin{equation}
J(\pi) =  E_{\pi} \left[ \sum_{t=0}^{T-1} \gamma r(s_t, a_t) \right]
\end{equation}
	
\section{Reactive Obstacle Avoidance Policies}

\begin{figure}
    \centering
    \includegraphics[width=0.95\textwidth]{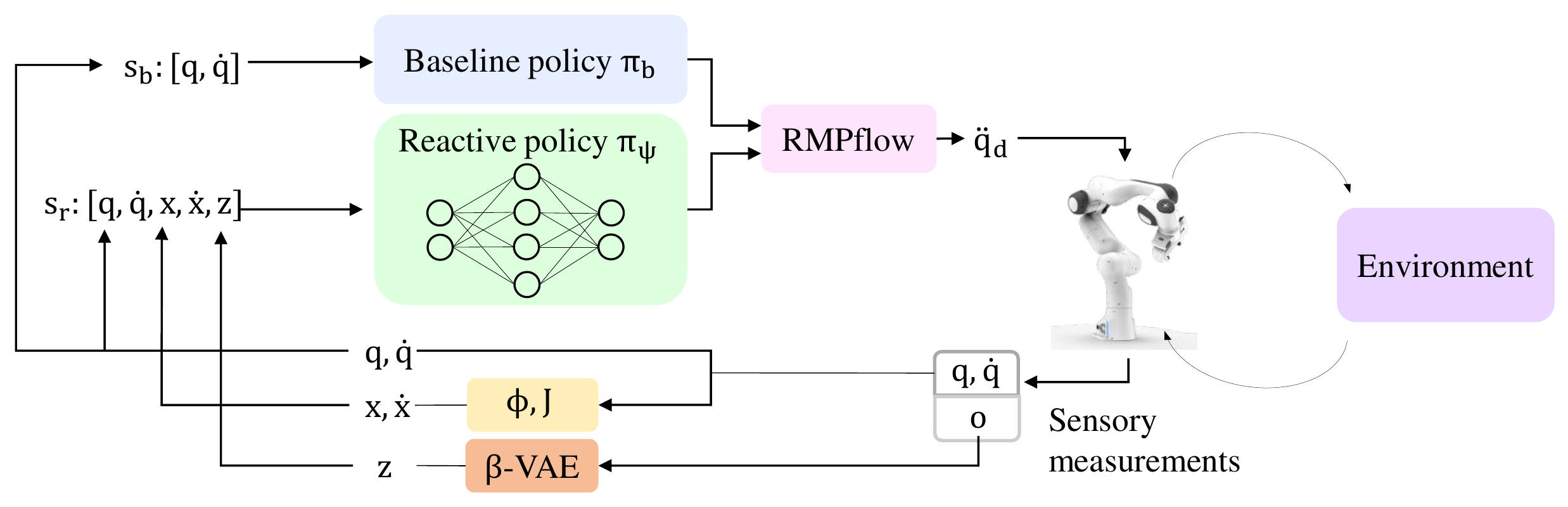}
\caption{System Overview: At a certain time step, the system receives a visual input $\textbf{o}$, joint angle measurements $\textbf{q}$ and joint velocities $\dot{\textbf{q}}$. The image $\textbf{o}$ is then encoded using a $\beta$-VAE encoder to produce the visual latent $\textbf{z}$. Using the robot kinematics $\phi$ and jacobian $\textbf{J}$, we obtain the end-effector position $\textbf{x}$ and velocity $\dot{\textbf{x}}$ from $\textbf{q}$ and $\dot{\textbf{q}}$ respectively. Subsequently $\textbf{q}$ and $\dot{\textbf{q}}$ are fed into a baseline policy to produce a user-defined behavior. Simultaneously, we feed all available information $s_r$ into a reactive policy $\pi_{\psi}$. The latter produces a reactive behavior dependent on the objects in the environment. Both outputs are then composed together based on the RMPflow framework, to produce a desired joint acceleration $\ddot{\textbf{q}}_d$. This acceleration is then fed into the robot controller. } 
    \label{fig:system}
\end{figure}

Our main goal is to connect perception and motion in a single framework. Specifically, we aim to find the relationship between the visual input and the structure of the robot's motion. Intuitively, having such a relationship would mean that we can map visual data to information about the structure of the robot's motion in its environment. The explicit way to achieve this goal is to learn a mapping from raw images to the parameters $\zeta$ of the corresponding Riemannian metric $\textbf{M}_{\zeta}(\textbf{x},\dot{\textbf{x}})$. However, due to the large number of parameters in $\zeta$, training such as mapping based on DRL would require a lot of samples. Instead, we consider an implicit method to achieve the same goal. Namely, we present a framework to learn vision-based reactive RMPs for obstacle avoidance. We design the action space and reward function in a way that encourages learning a similar but implicit representation of the vision-motion relationship. As modern machine learning methods depend on large datasets, robot learning in the real world can be time and resource expensive. In our method, we approach this problem from two different perspectives. First, we split the policy into two main parts. The first one is a hand-crafted \textbf{baseline policy}. The aim of this part is to guide the motion generation to achieve the task at hand. In our experiments, we consider goal-reaching tasks for which this policy can be based on simple principles such as point attractors. We then augment this strategy by a \textbf{reactive policy} which can correct the baseline behavior depending on the perceived situation. This policy reacts to obstacles based on its knowledge of the end-effector position, current robot configuration as well as visual input. Instead of learning it in an end-to-end fashion, we initially pre-train our visual
\textbf{latent model} based on data from random environment interactions. We then use the  obtained latent representations as input to the reactive policy. Both, the baseline and reactive policies are formulated as RMPs, which allows us to effectively combine them using RMPflow. The full system is shown in figure~\ref{fig:system}.

\subsection{Latent modeling}
\label{sec:latent}
For representation learning we first collect images from our environment by sampling actions based on brownian motion~\citep{uhlenbeck1930theory}. The primary information we hope to obtain in the latent representations would correspond to object poses, shape and geometric information of the objects in the environment. To achieve that, we reduce the episode length during the data collection phase. Since obstacle types and poses are random at different episodes, having shorter episodes would result in more episodes for the same amount of data and hence more diverse obstacles and obstacle poses in the collected samples. Once data is available, we train a $\beta$ Variational Autoencoder ($\beta$-VAE)~\citep{higgins2016beta} to obtain latent representations of our images. The objective function to train a $\beta$-VAE is the following:
\begin{equation}
    \mathcal{L}(\theta, \phi, \textbf{o}, \textbf{z}, \beta) = E_{q_{\phi}(\textbf{z}|\textbf{o})} [\log p_{\theta}(\textbf{o}|\textbf{z})] - \beta D_{KL}(q_{\phi}(\textbf{z}|\textbf{o}) || p(\textbf{z})) 
\end{equation}
where $\textbf{o}$ corresponds to images in our case, $\textbf{z}$ is the latent code, $\phi$ and $\theta$ are parameters of the encoder and decoder networks respectively, and $\beta$ is a regularization coefficient. Higher $\beta$ values help in learning more independent latent variables and hence improve disentanglement. It does so by forcing a stronger constraint on the latent bottleneck. Setting $\beta=1$ would then reduce this approach to a standard  VAE~\citep{kingma2013auto}. In our work, we use a $\beta$-scheduler based on~\citep{bowman2015generating}.

\subsection{Baseline policy}
\label{sec:baseline}
As previously mentioned, our primary task is goal reaching. For this purpose, we define a simple baseline RMP $\mathcal{R}_b= (\textbf{f}_b, \textbf{M}_b)_{\Omega_b}$ for goal reaching based on~\citep{ratliff2018riemannian, cheng2018rmpflow}, where ${\Omega_b}$ is a $3$-dimensional manifold in which the baseline policy is defined. Namely, for $\textbf{d}\in\Omega_b$, $\textbf{d}$ measures the distance between the end-effector and the goal along the $x,y$ and $z$ axes. Specifically, we use a GDS to define the RMP in $\Omega_{b}$. We first define the metric $\textbf{G}_b$ to be an identity matrix, which eliminates the curvature terms in the GDS and to $\textbf{M}_b:=\textbf{I}$ (based on the definitions in section~\ref{section:RMPflow}), where $\textbf{I}$ is the identity matrix. To ensure stability, we define the damping matrix as $\textbf{B}_b:=w\textbf{I}$,  where $w$ is a vector that consists of positive constants close in value to zero. Moreover, we define the gradient of the potential field $\nabla\Phi(\textbf{d}):=\textbf{d}$. Substituting these values in equation~\eqref{eq:RMP_force}, we obtain the GDS with the form
\begin{equation}
    \textbf{I}\ddot{\textbf{d}}=-\textbf{d}-w\textbf{I}\dot{\textbf{d}}.
\end{equation}
This would result in the forcing term $\textbf{f}_b=-\textbf{d}-w\textbf{I}\dot{\textbf{d}}$ according to equation \ref{eq:RMP_force}.
\begin{figure}
    \centering
    \includegraphics[width=\textwidth]{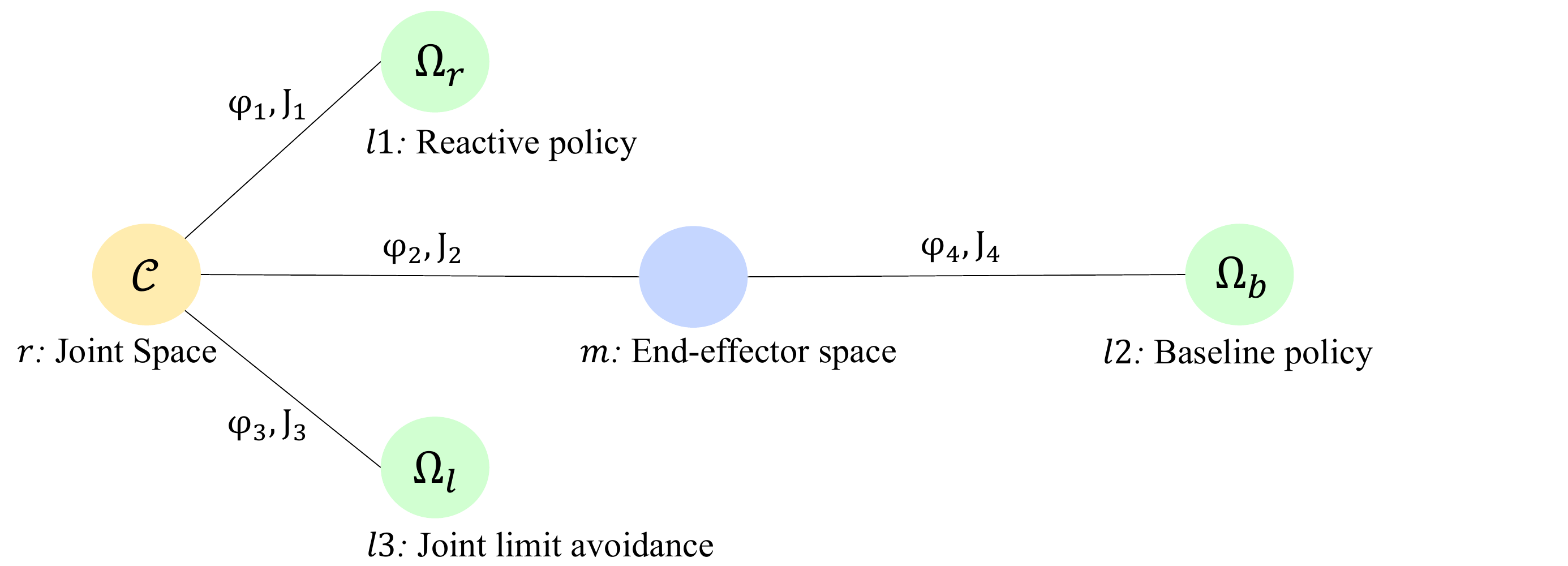}
    \caption{Structure of our RMP tree for combining multiple policies.}
    \label{fig:RMP_tree}
\end{figure}
\subsection{Reactive policy}
\label{sec:reactivep}
To enable the obstacle avoidance behavior, we define a reactive RMP $\mathcal{R}_r= (\textbf{f}_r, \textbf{M}_r)_{\Omega_r}$ in a $7$-dimensional manifold  $\Omega_r$. For $\textbf{d}_q\in\mathbb{R}^7$ in $\Omega_r$, $\textbf{d}_q$ measures the distance between the current joint configuration $\textbf{q}$ and the desired joint configuration $\textbf{q}_g$. We set the metric, the damping matrix and the potential field of this RMP similar to the ones in the previous section. According to equation~\ref{eq:RMP_force}, the resulting forcing term is: 
\begin{equation}
    \textbf{f}_r=-\textbf{d}_q-w\textbf{I}\dot{\textbf{d}}_q.
\end{equation}
We are then concerned in learning a mapping $\pi_{\psi}$ from an input state $s_r=[\textbf{q}, \dot{\textbf{q}}, \textbf{x}, \dot{\textbf{x}}, \textbf{z}]$ to an output action $a=\textbf{d}_q$, where $\textbf{x}$ corresponds to the end-effector position, and is obtained from $\textbf{q}$ using the kinematics $\phi_1$ of the manipulator, and $\dot{\textbf{x}}$ is obtained from $\dot{\textbf{q}}$ via the Jacobian $\textbf{J}_1$. As for $\textbf{z}$, it is the latent representation of a given image $\textbf{o}$ (as in section~\ref{sec:latent}). We model $\pi_{\psi}$ as a multilayer perceptron (MLP) with two hidden layers. It is then trained via residual RL~\citep{johannink2019residual} with the baseline policy from section~\ref{sec:baseline}. For this purpose we use the Twin Delayed Deep Deterministic policy gradient algorithm~\citep{fujimoto2018addressing}. Given the end-effector position $\textbf{x}$, the goal position $\textbf{x}_g$ and the action $a$ per step, the reward is defined as follows:
\begin{equation}
\label{eq:reward}
    r := r_{collide} + r_{goal} + r_{dist} + r_{control},
\end{equation}
where $r_{collide}$ is a negative reward of $-10$, which punishes the robot when colliding with an obstacle, $r_{goal}$ is a positive incentive of $5$ given only when the robot's end-effector reaches the goal, $r_{dist}=-1.6\|\textbf{x}-\textbf{x}_g\|+0.75$ is the distance reward which rises linearly when the distance between the end-effector and the goal decreases, and $r_{control}=-0.05\|a\|$ punishes large actions. This last term encourages the policy to diverge from the baseline policy only in cases of possible collision. The framework is not sensitive to changes in the values used in the reward function as long as the general relation between the terms is preserved. Namely, the distance reward is an important signal for training. When its magnitude is large, it leads to a local solution where the collision reward is neglected. Additionally, we aim to ensure that the episode rewards for goal-reaching and collision avoidance are not dominated by the control reward, hence the small weight 0.05. This combination of rewards encourages the resulting policy to move along the geodesic of the corresponding Riemannian Manifold. Furthermore, our MDP has a finite horizon with episodes ending at a specified maximum number of steps or whenever the robot collides with an obstacle. During early exploration stages, most episodes are unsuccessful and end up with collision. Consequently, only few successful trials are recorded. We use Prioritized Experience Replay~\citep{schaul2015prioritized} in order to use data from such trials more efficiently.

\subsection{Putting it all together}
In addition to the baseline and reactive policies defined in the previous sections, we use an additional RMP $\mathcal{R}_l=(\textbf{f}_l,\textbf{M}_l)_{\Omega_{l}}$ for joint limit avoidance similar to the one in~\citep{ratliff2018riemannian}. 
All three policies are then combined together based on the tree in figure~\ref{fig:RMP_tree}. The node $l_1$ uses the RMP $\mathcal{R}_r$ from section~\ref{sec:reactivep}, $l_2$ uses the $\mathcal{R}_b$ from section ~\ref{sec:baseline} and $l_3$ uses the joint limit avoidance RMP $\mathcal{R}_l$. Consequently, RMP-algebra is applied to compute the desired joint acceleration $\ddot{\textbf{q}}_d$. 

To ensure safety, we use impedance control with suitable collision handling algorithms guaranteeing contact force and torque thresholds, enabling quick collision detection and safe reaction~\cite{haddadin2008collision}. At the RL level, when a collision occurs during task execution, it is detected safely and early to prevent damage (based on~\cite{haddadin2008collision}), the episode terminates and the environment returns a negative reward as to inhibit such behavior in the future.

\section{Experiments}
\label{sec:result}
We conduct experiments to answer the following questions: (i) is our method capable of learning successful obstacle avoidance strategies? (ii) how does the baseline policy contribute to the overall sample-efficiency? (iii) is our approach capable of simultaneously avoiding multiple obstacles? 
\begin{figure}
    \centering
    \includegraphics[width=0.7\textwidth]{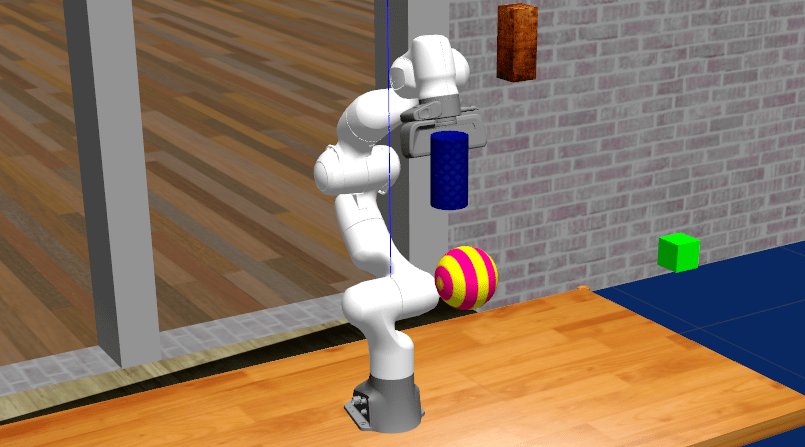}
    \caption{Illustration of our task setup in simulation.}
    \label{fig:env}
\end{figure}
\subsection{Setup}
All of our experiments share the same setup. Namely, we use a seven DoF Franka Emika Panda robot in a Gazebo simulation environment~\citep{koenig2004design}. We place a static RGB camera in front of the robot (green block in figure~\ref{fig:env}). We use three different obstacle shapes: cuboids, spheres and cylinders. Each obstacle has a distinct texture. In our experiments, we sample the object positions to be true obstacles with respect to the goal. The environment setup is illustrated in figure~\ref{fig:env}. Furthermore, we use an NVIDIA GeForce RTX 2080 GPU  to train our vision module, and reactive policy. For all of our experiments, we downsample the RGB images to $128 \times 128$. All of our results are averaged over $5$ training trials with different random seeds. We look at the success rate and the average episode return (AER). For improved visibility, we smoothen the AER plots using running means with a window size of $20$ episodes. For measuring the success rate, we label a trial as successful, when the end-effector reaches the desired goal with no collision and manages to stay there for $5$ seconds.

\subsection{Results}
\begin{figure}
\centering
   \begin{minipage}{0.48\linewidth}
   \centering
       \includegraphics[width=\textwidth]{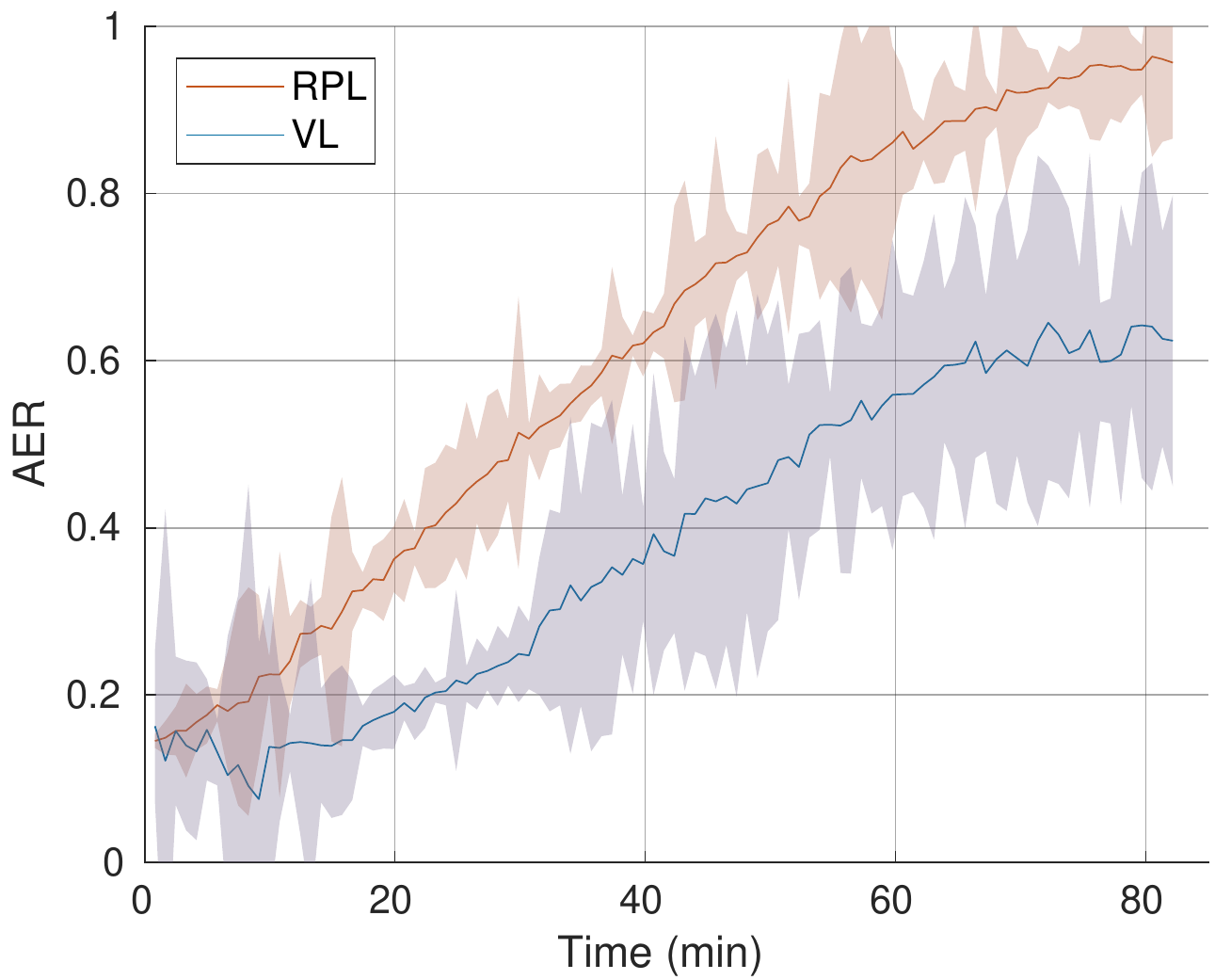}
     \end{minipage}%
   \begin{minipage}{0.5\linewidth}
   \centering
     \includegraphics[width=\textwidth]{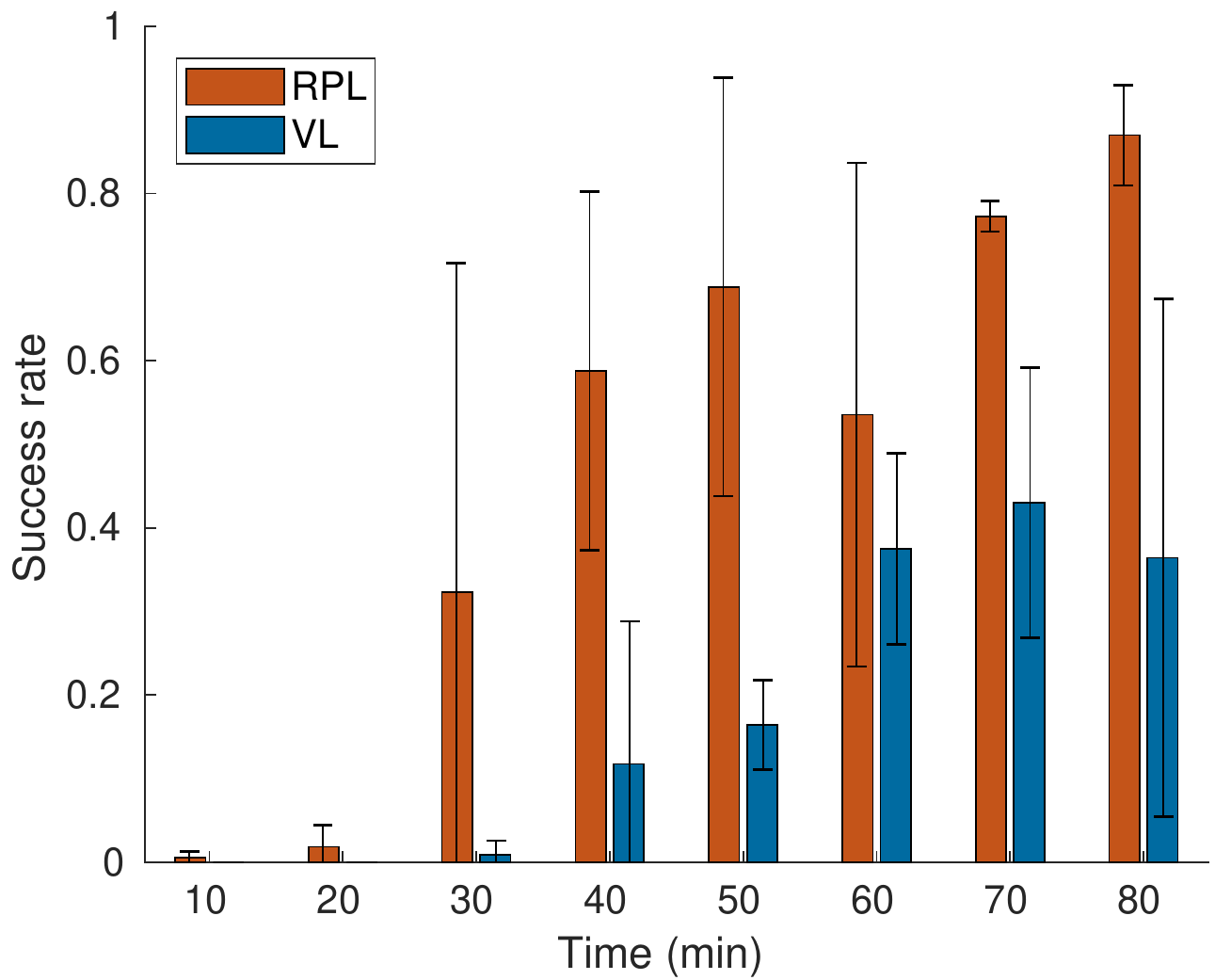}
     \end{minipage}
    \caption{\textbf{Experiment A}. Effect of the baseline policy on the single obstacle avoidance performance. (Left) Normalized average episode reward (AER) over time. (Right) Success rate over time.
    }
    \label{fig:ablation}
\end{figure}

\textbf{Experiment A.} In our first experiment, we want to evaluate our method on a single obstacle avoidance task. Additionally, we want to see the effect of the baseline policy on the overall performance. We refer to the case without a baseline policy as \textbf{Vanilla Learning (VL)} and the one with baseline policy as \textbf{Residual Policy Learning (RPL)}. VL would then correspond to a standard DRL approach for obstacle avoidance with the addition of the latent model and the RMP handling of $a=\textbf{d}_q$. In figure~\ref{fig:ablation}, we compare the average episode reward over time (left) and success rate (right) of both possibilities. After approximately $80$ min, both policies converge and the average reward as well as the success rates remain more or less the same. The total reward accumulated using RPL is substantially higher than that of VL. The latter manages to avoid the obstacle in few trials, however, it fails in most cases to also reach the goal, or even get close to it. This behavior can be shown by looking at the individual reward term from equation~\eqref{eq:reward}. We provide a plot of these values in the supplementary material. Furthermore, vanilla learning requires at least twice the amount of samples to reach the same success rate as our method. This is mainly due to two factors. First, for our use case,  VL attempts to learn both obstacle avoidance and goal reaching. However, in RPL, goal reaching is taken care of by the baseline policy. One could argue that providing a baseline policy limits the flexibility of the model, which usually leads to declined performance. However, our experiments show a different tendency. Even after convergence, our method achieves more than twice the success rate ($84\pm6\%$) than the vanilla method ($39\pm27\%$), which seems to often get stuck in a local solution (it reaches a lower average reward even after convergence). The second reason for this difference in sample-efficiency is exploration. In its early stages, our method already tries to reach the goal, as it is guided to do so by the baseline policy. While trying to do this, it also encounters the obstacles more frequently and receives negative reward. However, in the absence of a baseline, the exploration is random, and results mostly in low-reward trajectories for training. 

\begin{figure}
\centering
   \begin{minipage}{0.56\linewidth}
   \centering
       \includegraphics[width=\textwidth]{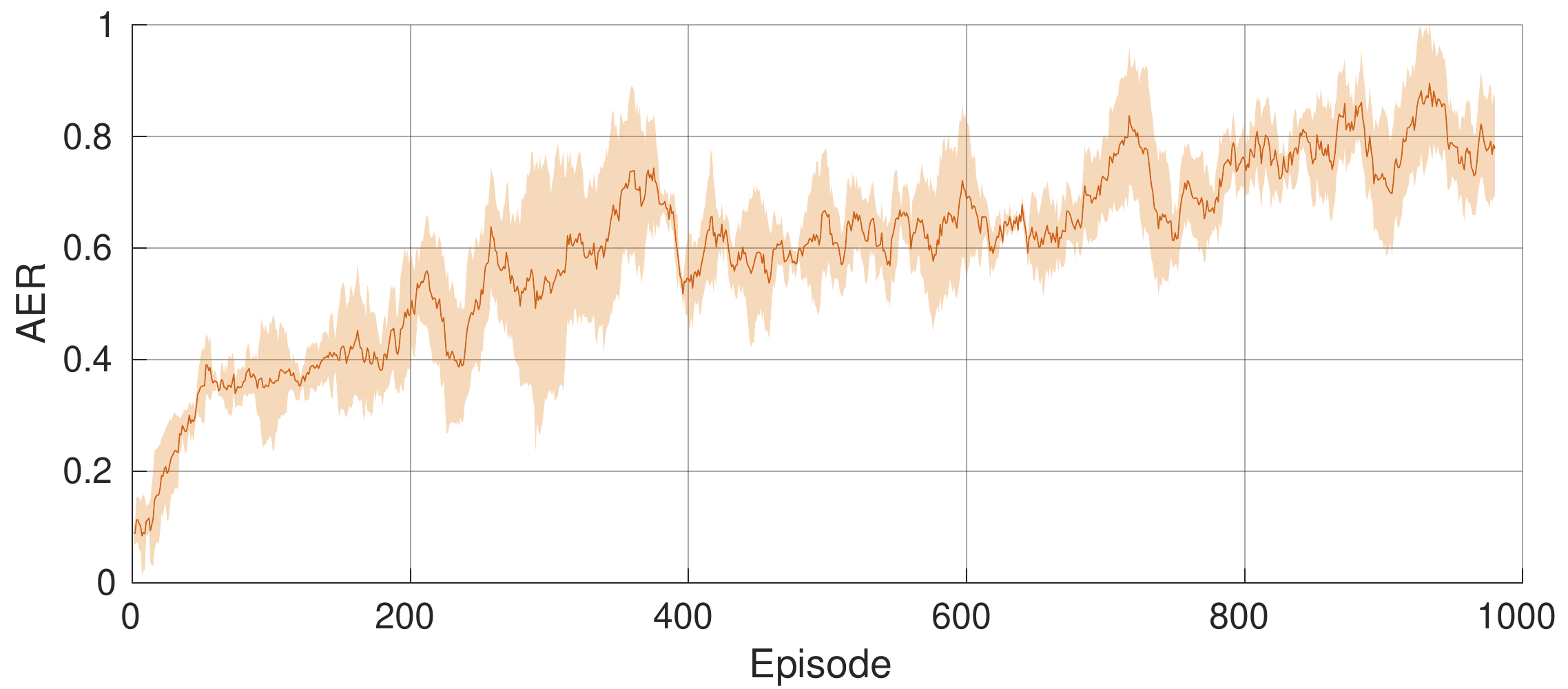}
     \end{minipage}%
   \begin{minipage}{0.44\linewidth}
   \centering
   \resizebox{\textwidth}{!}{
        \begin{tabular}{cccc}
        \hline
        \textbf{\# obstacles} & \textbf{Success} & \textbf{Near Goal} & \textbf{Failure} \\ \hline
        \textbf{1}                   & 66\%             & 30\%               & 4\% \\
        \textbf{2}                   & 54\%             & 44\%               & 2\% \\
        \textbf{3}                   & 72\%             & 22\%               & 6\%                               \\
                                       \hline
        \end{tabular}
      }
     \end{minipage}
    \caption{\textbf{Experiment B} Multiple obstacle avoidance. (Left) Normalized average episode reward (AER). (Right) Generalization of a policy trained for $3$ obstacles on scenarios with $1$ and $2$ obstacles. }
    \label{fig:multiob}
\end{figure}

\textbf{Experiment B.} In the second experiment, we aim to test the capability of our method to learn to avoid multiple obstacles simultaneously. The only difference is that we train the policy in an environment containing $3$ obstacles at all times (as shown in figure~\ref{fig:env}). As expected, the amount of interactions needed for this task to be learned is substantially higher than that of the single obstacle case. The policy converges after $10000$ episodes which is equivalent to $\sim3$ hours of data collection with a single robot. The success rate after convergence is $72\%$. We show the evolution of the AER over time in figure~\ref{fig:multiob} (left). Furthermore, we test the obtained policy on environments containing $1$ and $2$ obstacles to check whether the learned reactive behavior can generalize to different scenarios or just memorizes sequences of actions depending on the obstacles configuration. We evaluate our policy for $50$ trials for each scenario. We report \textbf{success} as previously defined, \textbf{near goal} reaching which corresponds to avoiding the obstacle and getting to the close proximity of the goal ($\sim7cm$) but not exactly reaching it, and \textbf{failure} which corresponds to all other cases. The results are shown in figure~\ref{fig:multiob} (right). Interestingly, even when only trained in $3$ obstacles environments, the obtained policy still manages to succeed in reaching the goal or its near proximity at a high rate, even when tested on environments with $1$ and $2$ obstacles. These results strongly support our claim for generalization.

\section{Conclusion}
\label{sec:conclusion}

    We propose a unified framework for learning vision-based obstacle avoidance strategies for robot manipulators. The presented approach treats perception and planning simultaneously and formulates the problem as a reinforcement learning problem in Riemannian Manifolds, successfully connecting perception and motion together. Our approach takes advantage of the latest advances in motion primitives as well as representation learning, to improve on the sample-efficiency and the safety of exploration aspects of learning. Based on that, we learn successful reactive policies, capable of avoiding both single and multiple obstacles. In addition, our experiments demonstrate the feasibility and sample-efficiency of our method. Finally, our framework is limited to static environments, we aim to address this issue in future work.




\acknowledgments{We greatly acknowledge the funding of this work by Vodafone AG, Microsoft Germany and the Alfried Krupp von Bohlen und Halbach Foundation. Please note that S. Haddadin has a potential conflict of interest as shareholder of Franka Emika GmbH.}


\bibliography{references}  

\appendix


\section{Implementation Details}
Previous work has shown that the performance of reinforcement learning algorithms is dependent on the implementation details~\citep{Henderson2017,rao2020howto}. Thus, we provide further details about our method's models and training procedures for both representation (section~\ref{sec:repl}) and reinforcement learning (section~\ref{sec:reil}) for the sake of reproducibility.

\subsection{Representation Learning}
\label{sec:repl}
Our $\beta$-VAE encoder is a convolutional neural network (CNN) with five convolutional layers having in order $6$, $32$, $64$, $128$ and $256$ channels. Each layer uses a rectified linear unit (ReLU) for activation and is followed by a batch normalization layer~\citep{ioffe2015batch}. The mean and log standard deviation layers of the latent are linear and are also followed by a batch normalization layer. The mean layer uses a hyperbolic tangent (Tanh) activation.

For training, we use the Adam optimizer~\citep{kingma2014adam} with a learning rate of $0.001$ and batches of size $128$.

\subsection{Reinforcement Learning}
\label{sec:reil}
TD3 is an actor-critic method~\citep{pmlrv80fujimoto18a}. We use an actor with two hidden layers, each containing 100 neurons. The activation function of all layers is Tanh. The critic has two hidden layers, with 500 neurons each. For the critic, we use ReLU activations for hidden layers and no activation for the output. We use Adam to optimize both sets of parameters. The training hyperparameters are listed in the table below.

\begin{table}[htb]

\centering
\begin{tabular}{cc}
\hline
\textbf{Hyperparameter}                           & \textbf{Value}\\ \hline
TD3 Policy noise                         & $0.2$                   \\
Max episode steps                       & $400$  \\         
Exploration noise                        & $0.5 \rightarrow 0.3$ \\
Memory Size                             & $300000$  \\        
Batch size                               & $64$ \\
Learning rate                           & $1e-3$ \\ \hline
\end{tabular}
\end{table}

\section{Simulation-Reality Gap}
We argue that the Simulation-to-Reality gap is not problematic for our experiments design. This is due to the nature of our task and goal, being concerned with pure geometric motion and not with dynamics, the latter being a lot harder to simulate. In our previous experiences, the main observed difference between simulation and the real-world, for vision-based motion generation tasks is in the noise distributions of images. This is due to mainly two factors. The first one is based on the lack of photo-realistic rendering in simulation environments. The second is due to the use of over-simplified environments with compact color backgrounds and minimum clutter. We compensate for this last factor by adding textures to the obstacles and by replicating the setting of common robot learning labs. Furthermore, VAE-based methods have been shown to capture real-world images well~\citep{van2016stable, lee2019making}. We use a $\beta$-VAE for our visual latent model, which would make the transition to real-world images only as expensive as the data collection process.

\section{Additional results}
In this section we show some additional results related to experiments A and B (section~\ref{sec:expa} and~\ref{sec:expb}) as well as to the latent model (section~\ref{sec:vae}).

\begin{figure}
\begin{tabular}{cc}
  \includegraphics[width=0.5\textwidth]{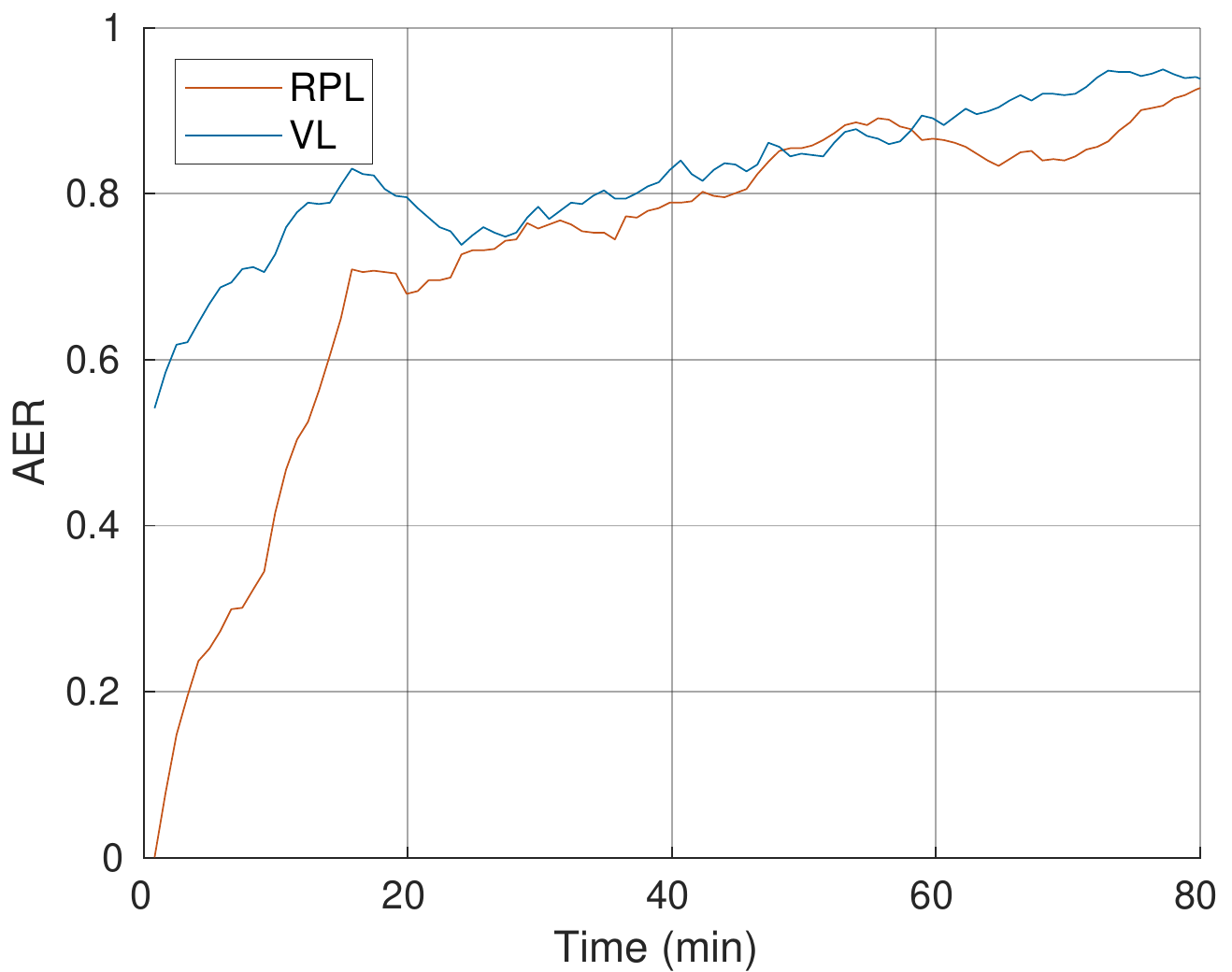} &   \includegraphics[width=0.5\textwidth]{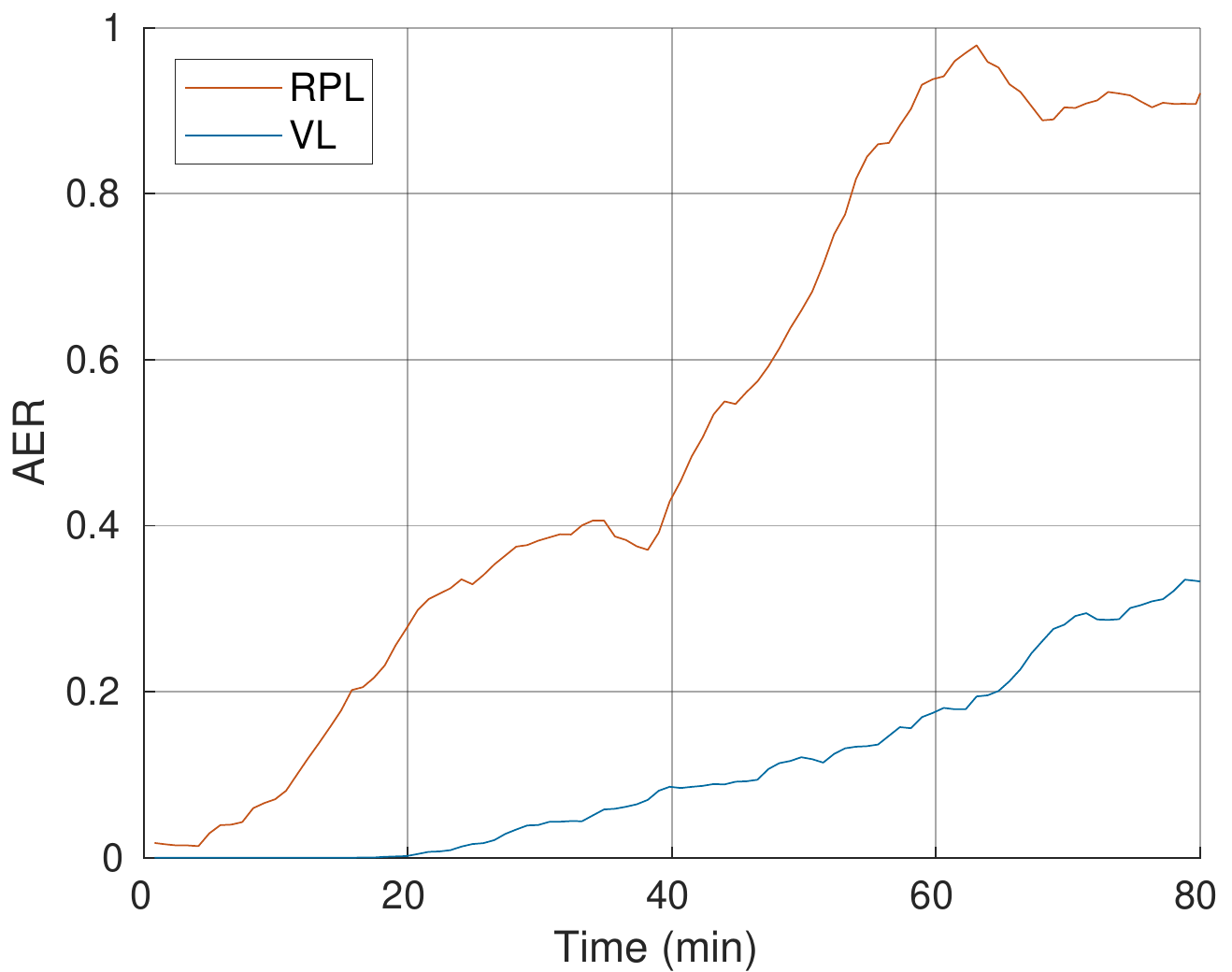} \\
(a) $r_{collide}$ & (b) $r_{goal}$ \\[6pt]
 \includegraphics[width=0.5\textwidth]{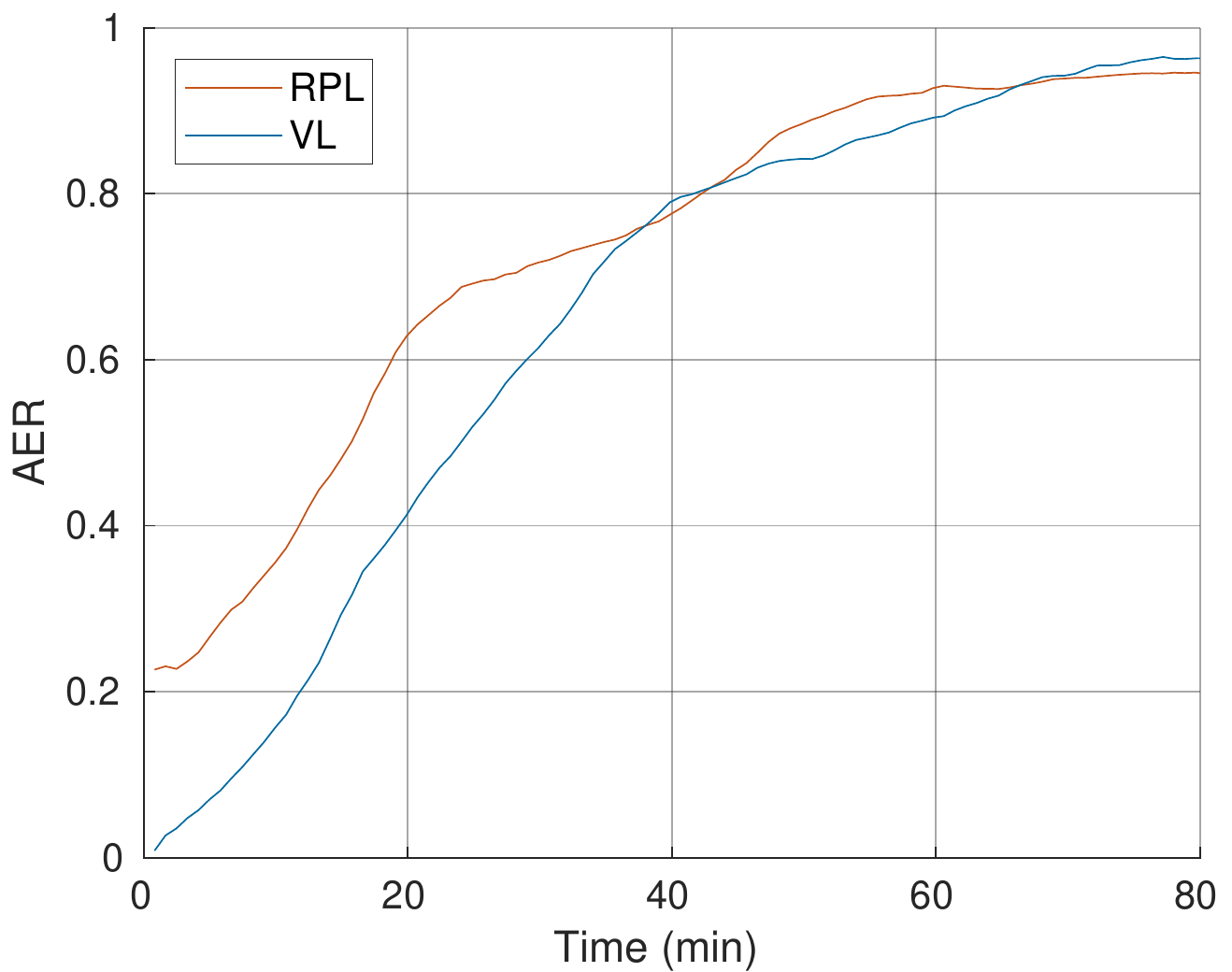} &   \includegraphics[width=0.5\textwidth]{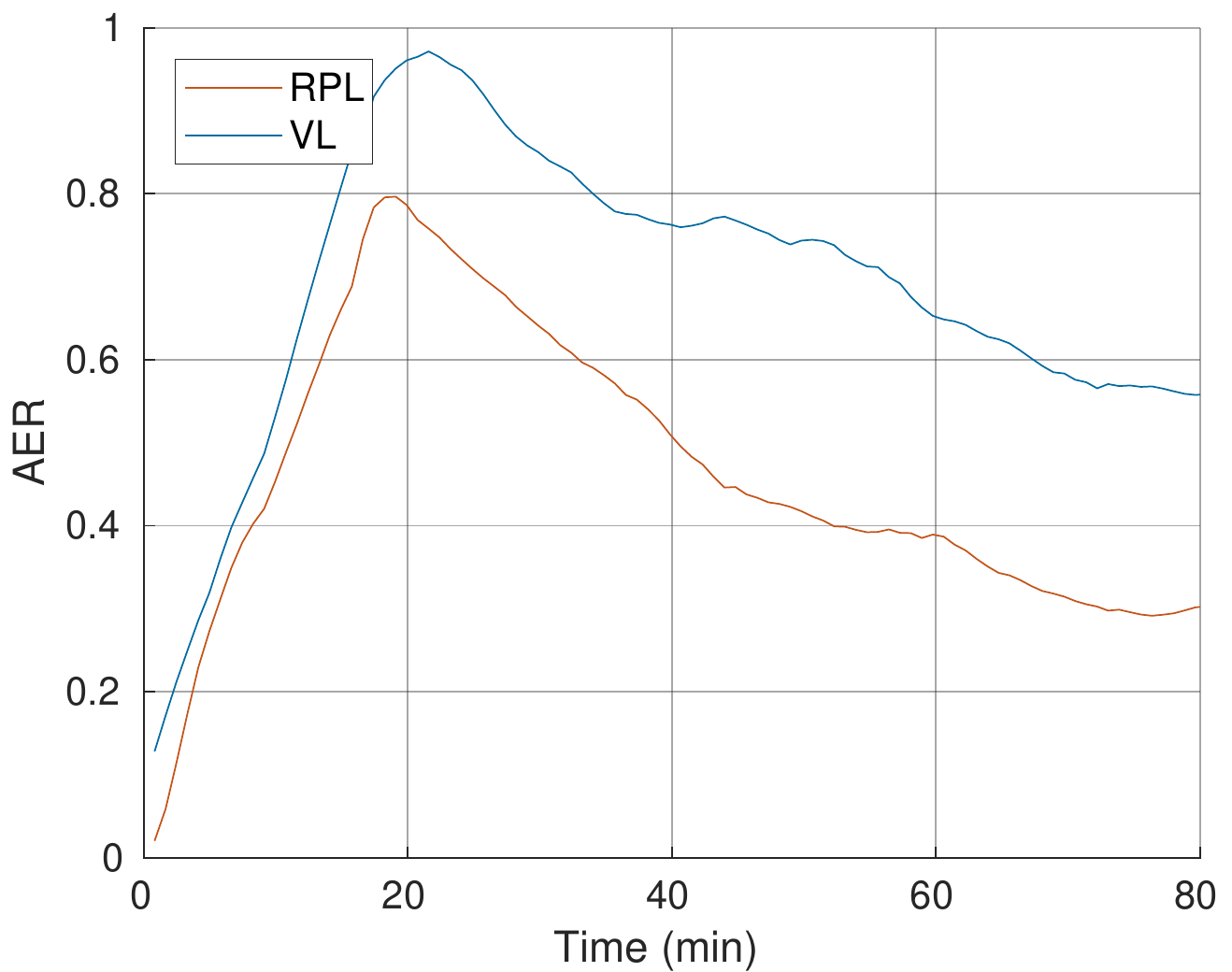} \\
(c) $r_{dist}$ & (d) $r_{ctrl}$ \\[6pt]
\end{tabular}
\caption{Individual rewards for the policy training with a baseline (RPL) and without (VL).}
\label{fig:rewards}
\end{figure}

\subsection{Experiment A}
\label{sec:expa}
In addition to the previously shown total reward plot, here we show plots of the individual reward terms: $r_{collide}$, $r_{goal}$, $r_{dist}$, $r_{control}$. These values are especially interesting when comparing the policy learning with the baseline and without it. As before, we refer to those cases as \textbf{Residual Policy Learning (RPL)}  and \textbf{Vanilla Learning (VL)} respectively. The results can be seen in figure~\ref{fig:rewards}. As expected the goal reward for RPL is substantially higher than that of VL at all times. At the beginning of training, RPL leads to more collisions with the obstacle as the baseline policy guides it towards the goal. Subsequently, RPL starts with more negative reward $r_{collide}$ than VL. However, it manages after training to reach a similar level as VL. In contrast, the latter depending mostly on random actions barely hits the obstacle at these stages as it's not even directed towards the goal. This can be seen in the $r_{goal}$ and $r_{dist}$ plots. As for the control reward $r_{ctrl}$, it behaves similarly for both methods. However, it gets higher for vanilla learning after a while. This could be explained by the following: RPL based exploration leads at all times to higher goal reward than VL. The latter, barely reaching the goal, prefers to take smaller actions to increase $r_{ctrl}$.

\subsection{Experiment B}
\label{sec:expb}
\begin{figure}
     \centering
     \includegraphics[width=0.8\textwidth]{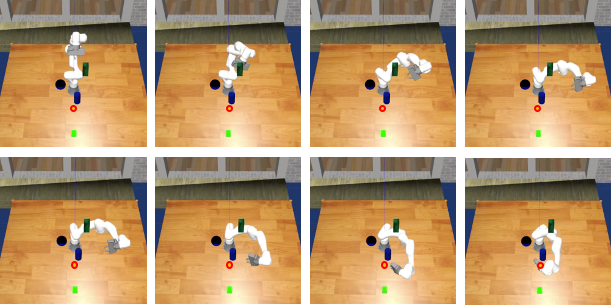}
     \caption{Multiple Obstacle Avoidance: Successful trial. The red circle is the goal.}
     \label{fig:expbsuc}
 \end{figure}

\begin{figure}
     \centering
     \includegraphics[width=0.8\textwidth]{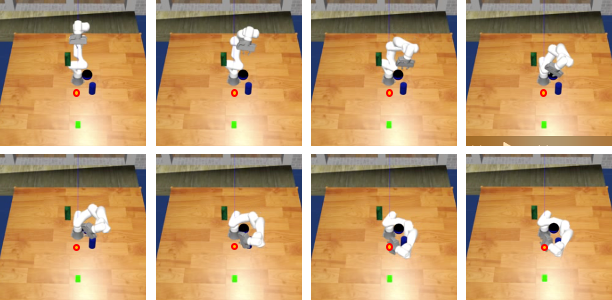}
     \caption{Multiple Obstacle Avoidance: Unsuccessful Trial. The red circle is the goal. The robot collides with the cylinder at the end of the execution.}
     \label{fig:expbunsuc}
 \end{figure}

In addition to the provided video here we show image sequences for our multiple obstacle avoidance results. We show a successful trial in figure~\ref{fig:expbsuc} and an unsuccessful trial in figure~\ref{fig:expbunsuc}.

\subsection{Experiment C}
\begin{figure}
    \centering
    \includegraphics[width=0.8\textwidth]{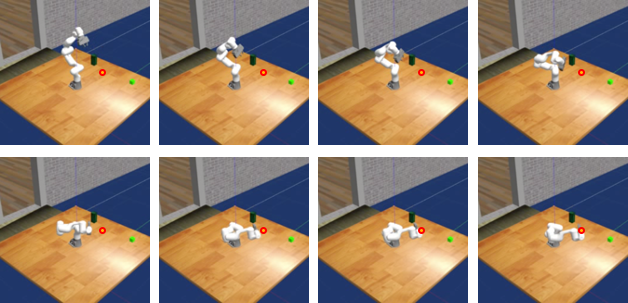}
    \caption{Single obstacle avoidance: cube. The red circle is the goal.}
    \label{fig:cube_trial}
\end{figure}

\begin{figure}
    \centering
    \includegraphics[width=0.8\textwidth]{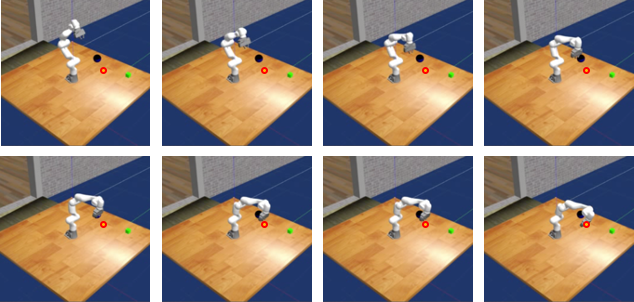}
    \caption{Single obstacle avoidance: sphere. The red circle is the goal.}
    \label{fig:sphere_trail}
\end{figure}

\begin{figure}
    \centering
    \includegraphics[width=0.8\textwidth]{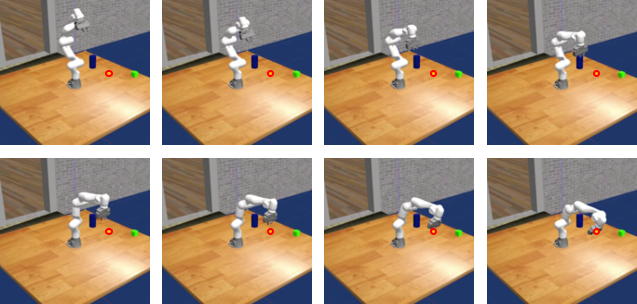}
    \caption{Single obstacle avoidance: cylinder. The red circle is the goal.}
    \label{fig:cylinder_trail}
\end{figure}

Although experiment B supports our claim of generalization, we conduct a further experiment to double-check our method's generalization ability. This experiment tests if the trained policy can generalize to unseen obstacles. We train our policy in single obstacle avoidance scenarios, with two types of obstacles: cuboid and sphere. After the training, we test the trained policy in scenarios with a cylinder obstacle. We evaluate the policy for $50$ trials in the scenarios containing the unseen obstacle. We report that the success rate is $72\%$, which is similar to the success rate when evaluating the policy using the previously-seen obstacles. Besides the training curve, we also provide an image sequence of the execution. Figure~\ref{fig:cube_trial} and~\ref{fig:sphere_trail} are evaluations of the policy on trained obstacles (cube and sphere). Figure~\ref{fig:cylinder_trail} shows a policy execution in the scenario with previously unseen obstacle (cylinder). The result of this experiment provides another evidence of the generalization ability of our method. 


\clearpage
\subsection{Sampling the latent}
\label{sec:vae}
\begin{figure}
\centering
\begin{tabular}{cccc}
  \includegraphics[width=0.15\textwidth]{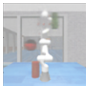} &  
  \includegraphics[width=0.15\textwidth]{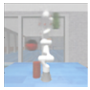} &  
  \includegraphics[width=0.15\textwidth]{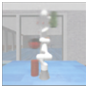} &  
  \includegraphics[width=0.15\textwidth]{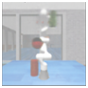} \\ 
  \includegraphics[width=0.15\textwidth]{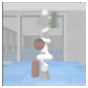} & 
  \includegraphics[width=0.15\textwidth]{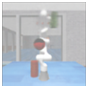} & 
  \includegraphics[width=0.15\textwidth]{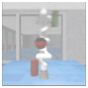} & 
  \includegraphics[width=0.15\textwidth]{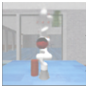} \\
  \includegraphics[width=0.15\textwidth]{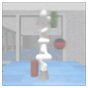} & 
  \includegraphics[width=0.15\textwidth]{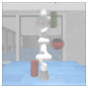} & 
  \includegraphics[width=0.15\textwidth]{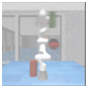} & 
  \includegraphics[width=0.15\textwidth]{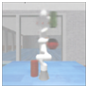} \\
  \includegraphics[width=0.15\textwidth]{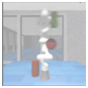} & 
  \includegraphics[width=0.15\textwidth]{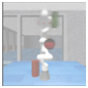} & 
  \includegraphics[width=0.15\textwidth]{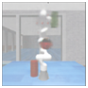} & 
  \includegraphics[width=0.15\textwidth]{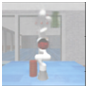} \\

\end{tabular}
\caption{Images Sampled from latent $z_0$ and $z_1$. First two rows: $z_1<0$ and $z_0 \in [-1,1]$ ; Last two rows, $z_1>0$ and $z_0 \in [-1,1]$. $z_0$ and $z_1$ control together the $y$-position of the sphere.}
\label{fig:latent1}
\end{figure}

 Here we illustrate images sampled from our VAE's latent as to show the importance of such variables to the obstacle avoidance task. The samples are shown in figure~\ref{fig:latent1}. Note that the sampled images are only blurry because of the down-sampling of the inputs images to the VAE.

\end{document}